\documentclass{article}

\usepackage[preprint]{neurips_2025}

\usepackage[utf8]{inputenc} 
\usepackage[T1]{fontenc}    
\usepackage{hyperref}       
\usepackage{url}            
\usepackage{booktabs}       
\usepackage{amsfonts}       
\usepackage{nicefrac}       
\usepackage{microtype}      
\usepackage{xcolor}         
\usepackage{wrapfig}        
\usepackage{adjustbox}
\usepackage{titlesec}

\usepackage{graphicx}	
\usepackage{amsmath}	
\usepackage{amssymb}	
\usepackage{booktabs}
\usepackage{times}
\usepackage{microtype}
\usepackage{epsfig}
\usepackage{caption}
\usepackage{float}
\usepackage{placeins}
\usepackage{color, colortbl}
\usepackage{stfloats}
\usepackage{enumitem}
\usepackage{tabularx}
\usepackage{xstring}
\usepackage{multirow}
\usepackage{xspace}
\usepackage{url}
\usepackage{subcaption}
\usepackage{xcolor}
\usepackage[hang,flushmargin]{footmisc}
\usepackage{makecell}
\usepackage{wrapfig, multirow}

\usepackage{booktabs}
\usepackage{multirow}
\usepackage{makecell}
\usepackage{graphicx} 
\usepackage[table]{xcolor} 
\usepackage{pifont} 
\usepackage[normalem]{ulem}

\newcommand{\cmark}{\textcolor{green!60!black}{\ding{51}}}
\newcommand{\xmark}{\textcolor{red!60!black}{\ding{55}}}

% ...

\newcommand{\R}[1]{{%
    \textbf{%
        \ifstrequal{#1}{1}{\textcolor{red}{R#1}}{%
        \ifstrequal{#1}{2}{\textcolor{blue}{R#1}}{%
        \ifstrequal{#1}{3}{\textcolor{magenta}{R#1}}{%
        \ifstrequal{#1}{4}{\textcolor{teal}{R#1}}{%
                           \textcolor{cyan}{R#1}%
        }}}}%
    }%
}}

\title{OpenEarth-Agent: From Tool Calling to Tool Creation for Open-Environment Earth Observation}

\author{
    Sijie Zhao\textsuperscript{1,2,\dag} \qquad
    Feng Liu\textsuperscript{2,3,\dag} \qquad
    Xueliang Zhang\textsuperscript{1*} \qquad
    Hao Chen\textsuperscript{2*} \qquad \\
    Xinyu Gu\textsuperscript{2} \qquad
    Zhe Jiang\textsuperscript{2} \qquad
    Fenghua Ling\textsuperscript{2} \qquad
    Ben Fei\textsuperscript{2} \qquad
    Wenlong Zhang\textsuperscript{2} \qquad \\
    Junjue Wang\textsuperscript{4} \qquad
    Weihao Xuan\textsuperscript{4} \qquad
    Pengfeng Xiao\textsuperscript{1} \qquad
    Naoto Yokoya\textsuperscript{4} \qquad
    Lei Bai\textsuperscript{2} \\ 
    \and 
    {\small
    \textsuperscript{1}Nanjing University} \quad
    \textsuperscript{2}Shanghai Artificial Intelligence Laboratory \\
    {\small
    \textsuperscript{3}Shanghai Jiao Tong University \quad
    \textsuperscript{4}The University of Tokyo} \\
    \vspace{1mm} \\
    {\small \dag~Equal contribution \quad *~Corresponding author}
}

\begin{document}
\maketitle

\begin{abstract}

Earth Observation (EO) is essential for perceiving dynamic land surface changes, yet deploying autonomous EO in open environments is severely hindered by the immense diversity of multi-source data and heterogeneous tasks. While remote sensing agents have emerged to streamline EO workflows, existing tool-calling agents are confined to closed environments. They rely on pre-defined tools and are restricted to narrow scope, limiting their generalization to the diverse data and tasks. To overcome these limitations, we introduce \textbf{OpenEarth-Agent}, the first tool-creation agent framework tailored for open-environment EO. Rather than calling predefined tools, OpenEarth-Agent employs adaptive workflow planning and tool creation to generate specialized tools tailored to unseen data and tasks. This adaptability is bolstered by an open-ended integration of multi-stage tools and cross-domain knowledge bases, enabling robust execution in the entire EO pipeline across multiple application domains. To comprehensively evaluate EO agents in open environments, we propose \textbf{OpenEarth-Bench}, a novel benchmark comprising 596 real-world, full-pipeline cases across seven application domains, explicitly designed to assess agents' adaptive planning and tool creation capabilities. Only essential pre-trained model tools are provided in this benchmark, devoid of any other predefined task-specific tools. Extensive experiments demonstrate that OpenEarth-Agent successfully masters full-pipeline EO across multiple domains in the open environment. Notably, on the cross-benchmark Earth-Bench \cite{earth_agent}, our tool-creating agent equipped with 6 essential pre-trained models achieves performance comparable to tool-calling agents relying on 104 specialized tools, and significantly outperforms them when provided with the complete toolset. In several cases, the created tools exhibit superior robustness to data anomalies compared to human-engineered counterparts, highlighting the potential of tool-creating agents for advanced EO. Code and Benchmark wiil be available at \href{https://github.com/walking-shadow/OpenEarth-Agent}{OpenEarth-Agent}.

\end{abstract}

\section{Introduction}

Earth Observation (EO) serves as a core paradigm for perceiving dynamic land surface changes and understanding human-earth systems, aiming to transform massive raw observation signals into actionable insights within open environments \cite{intro1}. The intrinsic nature of the open environment is fundamentally characterized by the diversity of EO data and tasks \cite{intro2}. Specifically, the data encompasses multi-modal observations (e.g., multispectral, Synthetic Aperture Radar (SAR)) and diverse derived products (e.g., land cover, wildfire) \cite{intro3_1,intro3_2}. Concurrently, EO tasks entail the diverse analysis of heterogeneous land surface elements, including urban, forests, and soils \cite{intro4}. In such an open environment, EO necessitates a long-horizon pipeline ranging from data preparation to geospatial analysis, and finds extensive applications across diverse domains, such as urban studies and agriculture \cite{intro5}. Consequently, this poses substantial challenges for autonomous EO in open environments.

\begin{figure}[htbp]
  \centering
  \includegraphics[width=\textwidth]{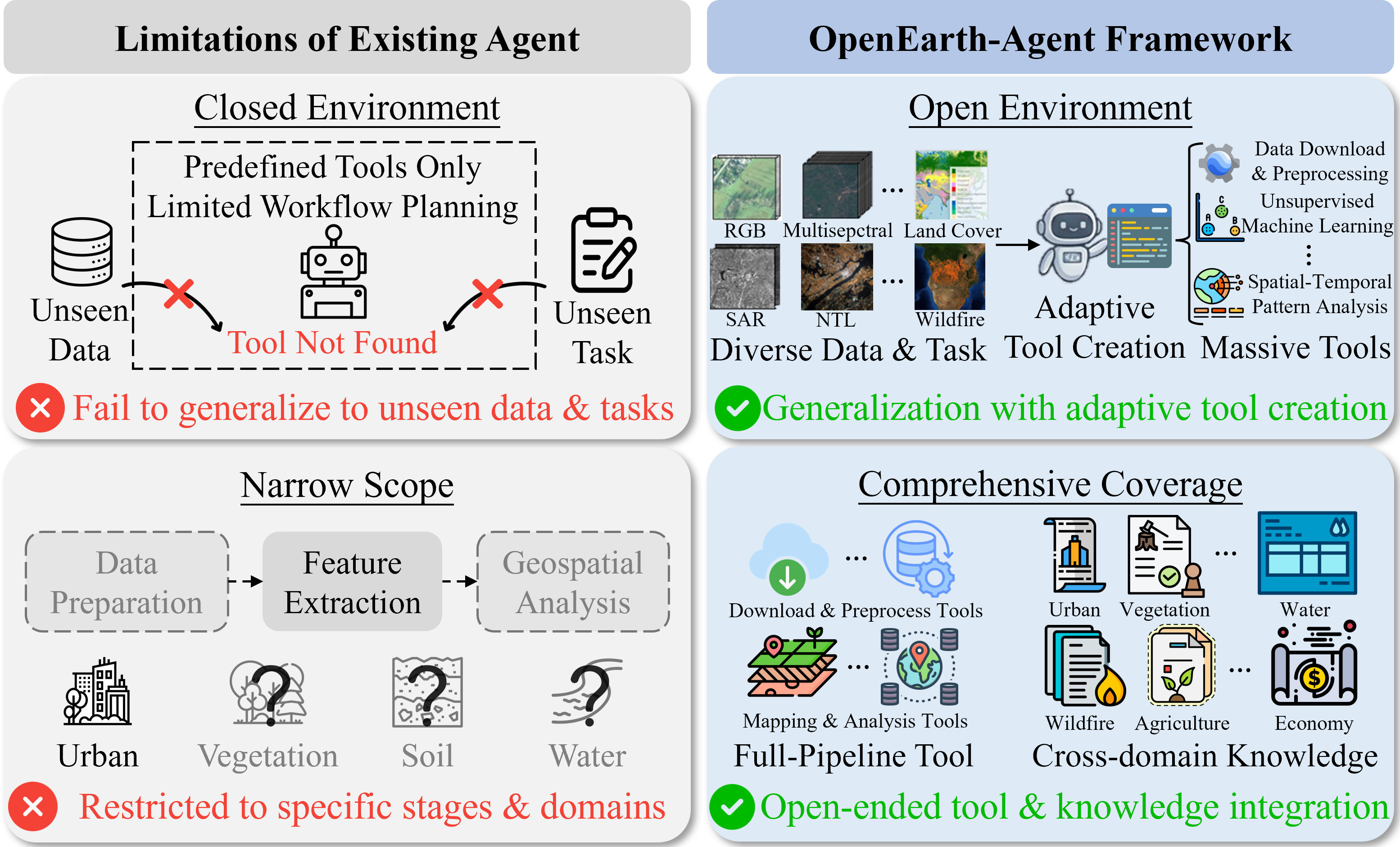}
  \caption{Overview of existing limitations and proposed solutions. Existing EO agents are failing to generalize to unseen data and tasks due to calling predefined tools, and are restricted to specific EO stages and application domains. In contrast, the proposed OpenEarth-Agent generalize to diverse data and tasks by creating adaptive tools, achieving full EO workflow and cross-domain coverage through open-ended tool and knowledge integration.
  }
  \label{fig:intro}
\end{figure}

The escalating complexity of EO has catalyzed the development of tool-calling agents within the remote sensing domain \cite{rsagent,rs_chatgpt,ringmo_agent}. By planning and calling predefined external tools to execute workflows, remote sensing agents decompose complex EO tasks into node-centric executions along a fixed toolchain, thereby facilitating autonomous and efficient EO \cite{thinkgeo,earth_agent,earthagent,cangling}. However, existing remote sensing agents predominantly operate within restricted, closed environments, focusing on the planning and invocation of predefined tools, as shown in Fig. \ref{fig:intro}. When deployed in expansive open environments, they exhibit two fundamental limitations: \textbf{1) Closed Nature of Tool Calling Limits Data and Task Generalization:} EO in open environments involves a high degree of data and task diversity, making it intractable to construct predefined tools for all potential scenarios. Nevertheless, current agents are restricted to calling a given set of tools and rely on these fixed tools as nodes for workflow planning. When a data or task in the open environment demands previously unseen functionality, the agent is unable to execute. \textbf{2) Restricted Operational Scopes Limit End-to-End Cross-Domain Generalization:} Open-environment EO encompasses a long pipeline and spans multiple cross domains. However, existing agents are typically restricted to a few specific stages and domains. Rather than autonomously acquiring and preprocessing data, they often read prepared datasets directly, executing only feature extraction or rudimentary analytical tasks. Furthermore, their cross-domain capabilities are limited to a narrow set of domain-specific cases, lacking the generalizability for EO across multiple domains.

To address these challenges, we propose \textbf{OpenEarth-Agent}, the first tool-creation agent framework tailored for open-environment EO, as shown in Fig. \ref{fig:intro}. OpenEarth-Agent tackles the aforementioned limitations through two core mechanisms: \textbf{(1) Adaptive Workflow Planning and Tool Creation:} To overcome the generalization bottleneck inherent in predefined tool calling, OpenEarth-Agent replaces static tool calling with a dynamic tool creation paradigm. Unlike conventional approaches that rely on simple code generation for tool creation, OpenEarth-Agent performs real-time data exploration and task logic planning to provide comprehensive and structured data and task contexts. Furthermore, upon tool creation and task execution, it leverages domain-specific geoscience knowledge for result verification and iteratively optimizes the synthesized tools. Specifically, confronted with diverse data and tasks, OpenEarth-Agent first synthesizes code-probing tools to perceive the metadata and spatio-temporal-modal characteristics of multi-source data in real time. This crucial grounding step ensures that the subsequently created tools are strictly calibrated to correctly process heterogeneous data. Conditioned on these data insights and task context, the agent then performs adaptive workflow planning, structuring the task execution into a Directed Acyclic Graph (DAG). Rather than calling fixed tools, OpenEarth-Agent dynamically creates adaptive tools explicitly tailored to the specific data and task requirements of each DAG node. Finally, through an iterative feedback mechanism, OpenEarth-Agent continuously refines these generated tools, fundamentally empowering it to seamlessly generalize to unseen data and novel tasks in open environments. \textbf{(2) Open-Ended Tool and Knowledge Integration:} To bolster the agent's capabilities across the entire EO pipeline and various application domains, OpenEarth-Agent incorporates extensive tool and knowledge integration. Departing from a rigid adherence to predefined workflows and static tool calling, OpenEarth-Agent leverages these resources as critical references to enable dynamic planning and real-time tool synthesis, adapting to heterogeneous data and complex tasks. Regarding tools, it integrates specialized functions and templates for multi-source data acquisition and preprocessing (e.g. multispectral, SAR, nighttime light), multi-paradigm feature extraction (e.g. index calculation, unsupervised clustering, pre-trained model calling), and multi-dimensional geospatial analysis (e.g. spatial pattern, temporal trend), comprising $66,736$ functional tools and $514$ execution templates. For knowledge integration, it encompasses $11,694$ static knowledge bases across domains like urban, agriculture, and soils, while concurrently supporting real-time online knowledge retrieval.

To comprehensively evaluate the performance of remote sensing agents in open environments, we construct \textbf{OpenEarth-Bench}, the first open-environment evaluation benchmark for EO, as shown in Fig. \ref{fig:OpenEarthbench}. Previous benchmarks typically conceptualize predefined tools as nodes, linking them to form cases within closed environments and assess the agent's tool planning and invocation capabilities. In contrast, OpenEarth-Bench evaluates the intrinsic ability to adaptively plan workflow and create tools without relying on predefined tools in open environments. OpenEarth-Bench comprises $596$ real-world application cases across $7$ application domains (e.g. urban, agriculture, vegetation). Each case encapsulates a full EO pipeline from data preparation and feature extraction to geospatial analysis. To ensure comprehensive coverage of diverse EO data and tasks, the benchmark includes the acquisition and preprocessing of multi-source observations and products; feature extraction utilizing statistical learning, machine learning, and deep learning methodologies; and in-depth geospatial analysis across multiple dimensions, such as temporal trend analysis, spatial pattern analysis, and spatio-temporal coupling analysis.

To assess the efficacy of OpenEarth-Agent in open environments, we evaluate its performance powered by various Large Language Models (LLMs) on OpenEarth-Bench. Furthermore, to rigorously evaluate its workflow planning and tool creation capabilities, we conduct cross-benchmark experiments on Earth-Bench. To isolate the effects of knowledge and tool integration, we configure OpenEarth-Agent without any external knowledge or extraneous tools in this benchmark. Under this controlled setting, we compare OpenEarth-Agent equipped with either only $6$ essential pre-trained model tools or all $104$ tools from the benchmark, against other agents that integrate and invoke the complete toolset.

Overall, the primary contributions of this work are summarized as follows:
\begin{itemize}
    \item We propose OpenEarth-Agent, the first remote sensing agent architecture designed for open environments. By performing adaptive workflow planning and tool creation conditioned on data and task contexts, OpenEarth-Agent accommodates diverse EO data and tasks. Through multi-stage tool and cross-domain knowledge integration, it effectively executes full-pipeline EO across multiple application domains.
    \item We construct OpenEarth-Bench, the first remote sensing agent benchmark oriented towards open environments. Comprising $596$ full-pipeline cases sourced from real-world applications across various domains, it serves as a robust platform for evaluating agent performance in open environments.
    \item Extensive experiments on OpenEarth-Bench and cross-benchmark evaluations on Earth-Bench validate the effectiveness of OpenEarth-Agent. Notably, on Earth-Bench, OpenEarth-Agent demonstrates the ability to create functionally equivalent professional tools, and in certain instances, tools with superior data adaptability. Operating with merely $6$ integrated tools, it achieves performance comparable to agents utilizing full tool calling, and significantly outperforms existing agents when provided with the complete toolset.
\end{itemize}

\section{Related Works}

\subsection{Remote Sensing Agents for Earth Observation}
Early applications of LLMs in EO predominantly relied on developing domain-specific remote sensing encoders for end-to-end multimodal alignment \cite{related_1_1, related_1_2}. However, when confronted with heterogeneous remote sensing data and tasks, this paradigm is highly expensive to scale and relies heavily on exhaustive fine-tuning \cite{related_2_1}. Recent research has progressively shifted towards constructing tool-augmented agents. By calling and orchestrating predefined external tools, these agents execute designated workflows, thereby transforming complex EO tasks into node-based executions within a fixed toolchain \cite{thinkgeo,cangling,earth_agent,earthagent,geoevolver}. Nevertheless, most existing remote sensing agents are confined to idealized, closed-world testing environments. They depend on static, predefined toolsets for restricted planning and are typically constrained to isolated procedural stages within limited domains. Consequently, they struggle to address the cross-domain and full-pipeline demands of real-world EO. Thus, there is a pressing need to develop open-environment EO agents capable of autonomous operation across diverse domains and entire workflows.

\subsection{Remote Sensing Agent Benchmarks}
Early remote sensing multimodal benchmarks integrating LLMs primarily evaluated the models' fundamental perception and reasoning capabilities \cite{related_3_1, related_3_2, related_3_3}. Recent efforts have begun to construct benchmarks specifically for remote sensing agents, shifting the focus towards evaluating tool planning and invocation capabilities following the integration of specialized tools \cite{thinkgeo,cangling,earth_agent,earthagent}. However, these existing benchmarks are generally designed around closed environments, requiring agents to orchestrate workflows strictly within a predefined set of tools. Consequently, such benchmarks remain restricted to a limited number of pre-configured scenarios, lacking a comprehensive evaluation of full EO pipelines and cross-domain applications in open environments. There is currently an urgent need to establish benchmarks that rigorously assess the real-world capabilities of EO agents in open environments.

\section{Methodology}

\begin{figure}[htbp]
  \centering
  \includegraphics[width=\textwidth]{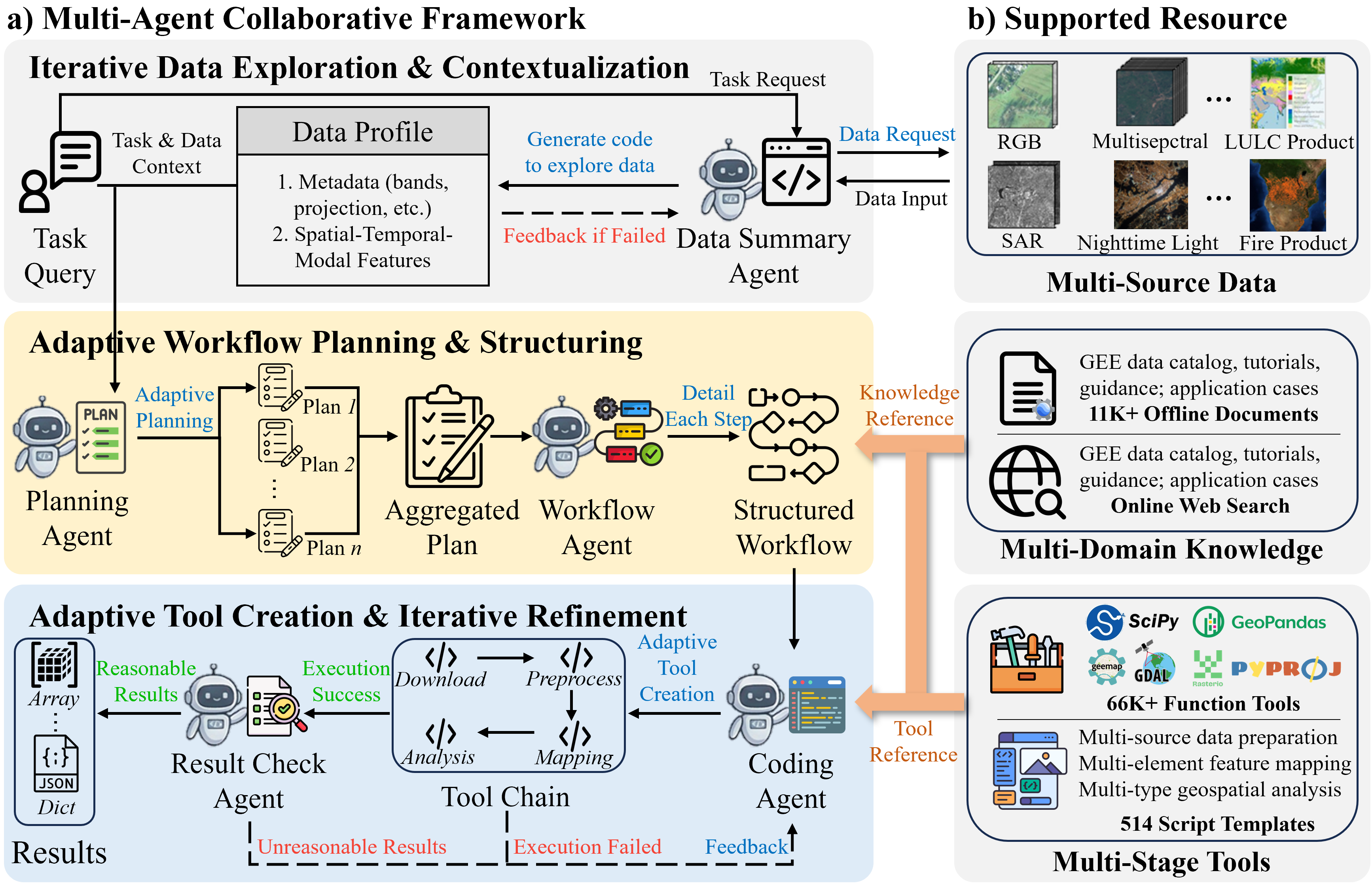}
  \caption{Overview of the OpenEarth-Agent. The OpenEarth-Agent performs real-time perception of multi-source EO data, followed by adaptive workflow planning and structuring based on the task and data context. It then proceeds with adaptive tool creation and execution, incorporating real-time inspection and feedback on the execution process and outcomes to generate reasonable results.
  }
  \label{fig:method}
\end{figure}

\subsection{Multi-Agent Architecture}
To address the high diversity and complexity of EO data and tasks in open environments, we propose OpenEarth-Agent, a multi-agent collaborative framework, as shown in Fig. \ref{fig:method}. Unlike existing tool-calling agents that rely on predefined toolsets and are restricted to narrow scope, our architecture aims to achieve an autonomous closed-loop for the entire EO pipeline across multiple domains through adaptive workflow planning and tool creation. OpenEarth-Agent operates collaboratively via five core agents: Data Summary Agent, Planning Agent, Workflow Agent, Coding Agent, and Checking Agent. The overall operational logic can be divided into the following three core stages:

\textit{1) Iterative Data Exploration and Contextualization:} In open environments, flexibly adapting to multi-source remote sensing data is a prerequisite for autonomous EO. Given the natural language task instruction $\mathcal{Q}$, the Data Summary Agent first retrieves and acquires the relevant raw data. Rather than relying on static parsers, the agent dynamically generates and executes specialized data-probing tools to extract a comprehensive data profile $\mathcal{D}$. This profiling captures both essential metadata (e.g., file paths, dimensions, spectral bands, and spatial projections) and complex spatial-temporal-modal features. Crucially, this exploration process is strictly iterative: if a probing tool encounters an execution error due to underlying data irregularities, the agent autonomously analyzes the traceback, refines the script, and re-executes the tool until the data is successfully parsed. Finally, the agent distills the extracted characteristics into a rich, structured data context, establishing an accurate and reliable informational foundation for subsequent workflow planning and tool creation.

\textit{2) Adaptive Workflow Planning and Structuring:} After data contextualization, the Planning Agent receives the natural language task instruction $\mathcal{Q}$, and the aforementioned data profile $\mathcal{D}$, leveraging the integrated cross-domain expert knowledge base $\mathcal{K}$ and multi-stage tools $\mathcal{T}$ to generate a reasonable plan. This agent is responsible for decoupling the macroscopic task into a sequence of subtasks with a clear hierarchical structure and execution order. To effectively mitigate the planning hallucination problem common in LLMs, we introduce a multiple-plan aggregation mechanism. The Planning Agent first generates a set of $n$ candidate execution blueprints, denoted as $\mathcal{P} = \{p_1, p_2, \dots, p_n\}$. It then merges execution paths with similar semantics and prioritizes them based on dimensions such as data availability and logical rigor, ultimately outputting the aggregated optimal task plan $p^*$. This process can be formalized as:
\begin{equation}
    p^* = \mathcal{A}(\mathcal{P}), \quad \mathcal{P} = \mathcal{G}(\mathcal{Q}, \mathcal{D}, \mathcal{K}, \mathcal{T})
\end{equation}
where $\mathcal{G}$ denotes the candidate plan generation function based on the task and data context, and $\mathcal{A}$ represents the plan aggregation and ranking function.

To bridge the semantic gap between macroscopic task planning and microscopic code generation, the Workflow Agent transforms the optimal plan $p^*$ into a structured computational workflow. Specifically, this agent decomposes the overall plan into multiple independent execution script nodes with well-defined functions. It strictly regulates the data flow dependencies between nodes and explicitly defines the input/output interface information and key execution parameters for each script node. This process instantiates the natural language planning blueprint into a logically rigorous DAG, providing a precise standard reference for subsequent code generation.

\textit{3) Adaptive Tool Creation and Iterative Refinement:} Guided by the structured workflow, the system proceeds to the tool creation and execution phase. Following the topological dependency order of the workflow, the Coding Agent sequentially generates execution scripts for each node. Unlike existing methods that retrieve and invoke tools from a limited toolset, the Coding Agent fully leverages the code generation capabilities of LLMs, combined with the integrated professional toolsets and knowledge base, to create adaptive and specialized tools within open environments. The agent advances to the tool creation for the next node only after the script of the preceding node has been successfully executed.

To ensure the robustness of the system in open environments, the Checking Agent constructs a feedback-driven validation closed loop. As the Coding Agent executes each script, the Checking Agent performs runtime diagnostics to determine whether the code runs successfully and utilizes geoscientific rules to verify the validity of the results (e.g., checking if saved results consist entirely of invalid values or if the calculated remote sensing indices fall outside logical numerical ranges). Once an anomaly is detected, the diagnostic feedback is transmitted back to the Coding Agent, triggering parameter correction and code refactoring mechanisms to initiate a new round of tool creation and execution. Letting the code output at the $k$-th round be $O^{(k)}$, this autonomous error-correction closed loop can be modeled as:
\begin{equation}
    O^{(k+1)} = \mathcal{E}\left(\mathcal{M}\left(\mathcal{C}^{(k)}, \mathcal{F}^{(k)}\right), \mathcal{D}\right), \quad \text{s.t.} \quad \mathcal{V}(O^{(k+1)}) = 1
\end{equation}
where $\mathcal{C}^{(k)}$ is the code generated in the $k$-th round, $\mathcal{F}^{(k)}$ is the anomaly feedback information provided by the Checking Agent, $\mathcal{M}$ denotes the code modifier, $\mathcal{E}$ represents the code executor, and $\mathcal{V}$ is the validation function used to determine whether the execution result meets the convergence criteria. This iterative feedback mechanism significantly enhances the robustness and scientific rigor of OpenEarth-Agent in handling complex tasks.

\subsection{Full-Pipeline Tool Integration}
To ensure that OpenEarth-Agent can reasonably create tools throughout the entire EO pipeline in open environments, we design a highly flexible tool integration paradigm. We construct the tool integration as a two-tier architecture comprising standardized software packages and an expert-level code script library, serving as the reference for tool creation.

At the standardized software package level, we integrate 66,736 function tools from 11 Python libraries across the domains of remote sensing, geocomputation, and general data science. This allows the agents to directly import and freely combine them within a standard Python environment to handle diverse foundational tasks. At the expert-level code script library level, we construct a modular repository validated by domain experts, covering complete or partial workflows for remote sensing applications, comprising 514 reference scripts. To enable autonomous invocation by the agents, we introduce a retrieval mechanism based on semantic similarity. First, an LLM generates a functional description for each script, which is then transformed into high-dimensional vectors via an embedding model to build a retrieval database. When an agent faces a new task, the system translates its requirements into natural language descriptions and performs vectorized matching. This retrieves the most relevant script pairs as high-quality reference examples and code templates, thereby significantly enhancing the accuracy and robustness of the agent's tool creation in remote sensing tasks.

Based on the aforementioned two-tier architecture, we conduct deep integration of specialized tools across three core stages of EO:

\textit{1) Multi-Source Data Acquisition and Preprocessing:} Facing complex data sources in open environments, OpenEarth-Agent integrates a comprehensive toolchain covering cloud-based acquisition and local preprocessing. For cloud-based data acquisition, it integrates interaction codes and application paradigms from the Google Earth Engine (GEE) platform. This enables the agent to dynamically acquire multi-modal raw observation data (e.g. multispectral data, nighttime light data and supports the direct invocation of global high-level remote sensing products (such as, land cover products, and wildfire products). For local data preprocessing, it comprehensively integrates function tools and operational templates from core spatial processing libraries like \texttt{gdal}, \texttt{rasterio}, \texttt{geopandas}, and \texttt{pyproj}, providing standardized code references for preprocessing operations such as resampling, projection transformation of multi-source data.

\textit{2) Multi-Paradigm Feature Extraction:} To support the efficient transformation from massive pixels to element features, OpenEarth-Agent integrates multi-level feature extraction tools ranging from statistical learning, machine learning to deep learning. At the statistical and machine learning level, it integrates powerful general-purpose algorithm libraries such as \texttt{scikit-learn} and \texttt{scikit-image}, enabling the agent to rapidly construct feature extraction models. At the deep learning level, it incorporates pre-trained model tools for tasks like building semantic segmentation and multi-temporal change detection.

\textit{3) Multi-Dimensional Geospatial Analysis:} To transform feature extraction results into a deep cognitive understanding of element states, OpenEarth-Agent integrates rich geostatistical analysis tools centered around three dimensions (temporal trends, spatial patterns, and spatiotemporal coupling). For time series analysis, it integrates libraries like \texttt{statsmodels} and \texttt{SciPy} to provide advanced statistical tool templates, such as autoregressive analysis and time series decomposition, to mine the long-term evolutionary patterns of element features. For spatial pattern and spatiotemporal coupling analysis, it deeply integrates the PySAL ecosystem, offering reference scripts for analysis like spatial autocorrelation analysis, hotspot analysis.

\subsection{Multi-Domain Knowledge Integration}
To address the complex and diverse EO demands in open environments, OpenEarth-Agent integrates knowledge from multiple remote sensing application domains, aiming to provide expert-level prior knowledge guidance for workflow planning and tool creation. Specifically, the integrated knowledge encompasses the GEE Data Catalog, tutorial documents, community documentation, and domain-specific knowledge extensively covering various remote sensing intersections, totaling 11,694 documents. Regarding the integration mechanism, we adopt a dual-branch knowledge retrieval architecture combining offline and online strategies:

\textit{1) Offline Knowledge Base:} For multi-source heterogeneous GEE documentation and various remote sensing application cases, we segment massive long texts into independent semantic chunks and vectorize each chunk to construct a dedicated document vector database. This database provides a natural language query interface. During the workflow planning and tool creation, they can initiate queries in natural language, and the system will precisely recall the highly relevant textual chunks.

\textit{2) Online Retrieval Tool:} Given the rapid iterations of remote sensing data and analytical methods in the real world, we equip the knowledge base with an online web search tool. When agents encounter knowledge blind spots, they can autonomously invoke this tool for real-time web retrieval. This allows OpenEarth-Agent to continuously acquire the latest academic literature, open-source code repositories, and updated API states at any time, ensuring the real-time nature of the system's knowledge and the continuous expansion of its boundaries.

\section{OpenEarth-Bench}

\subsection{Overview of OpenEarth-Bench}

\begin{figure}[htbp]
  \centering
  \includegraphics[width=\textwidth]{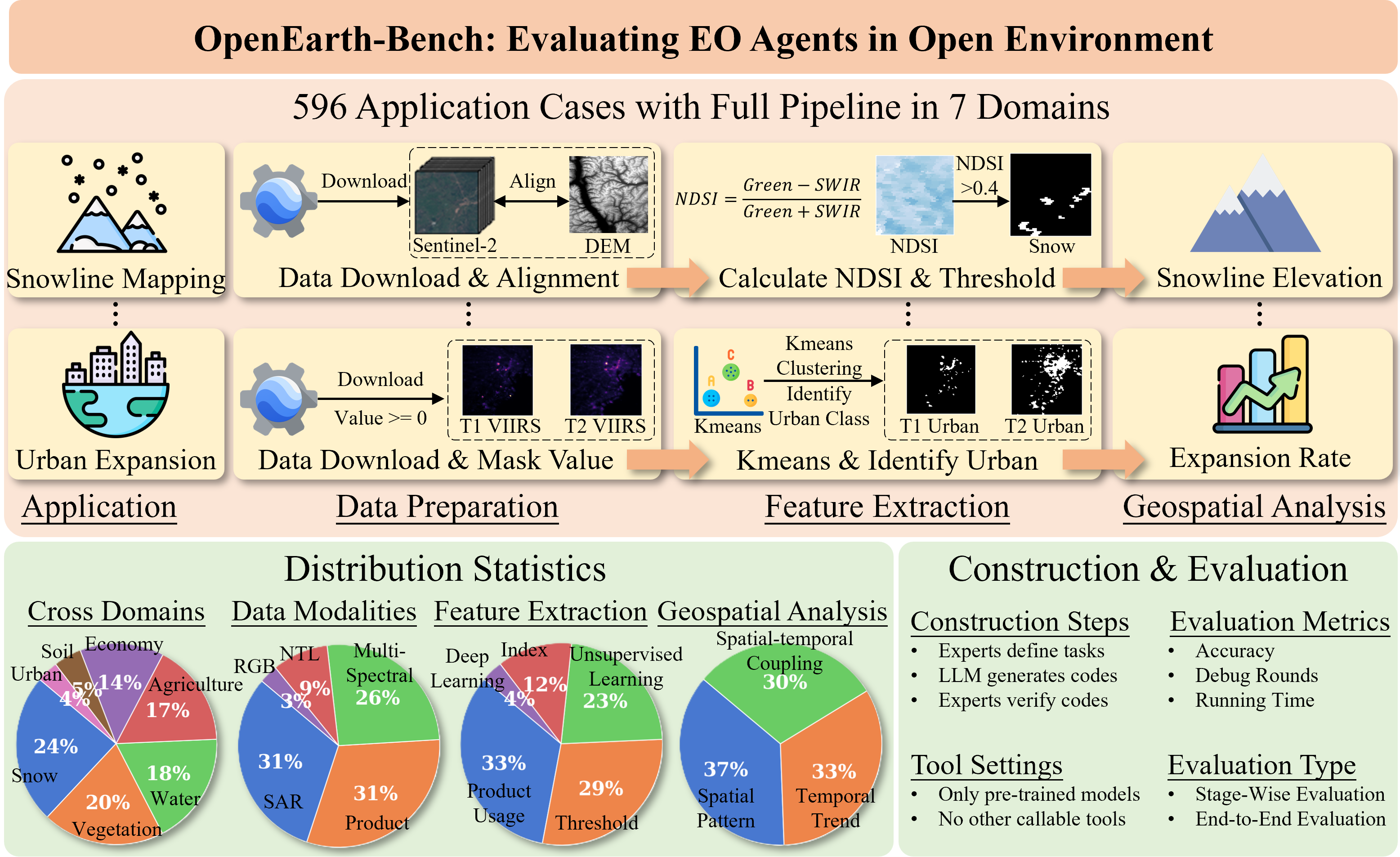}
  \caption{Overview of the OpenEarth-Bench. This benchmark comprises 596 real-world application cases spanning seven domains. Each case encompasses the entire workflow, ranging from data preparation and feature extraction to geospatial analysis.
  }
  \label{fig:OpenEarthbench}
\end{figure}

To comprehensively evaluate the autonomous EO capabilities of agents in open environments, we construct OpenEarth-Bench, the first open-domain evaluation benchmark for remote sensing agents. Unlike existing benchmarks that test the tool planning and invocation capabilities of remote sensing agents in closed environments, OpenEarth-Bench is designed to evaluate the adaptive workflow planning and tool creation capabilities of agents in open environments across full EO pipelines and diverse application domains, as shown in Fig. \ref{fig:OpenEarthbench}.

To systematically examine the comprehensive performance of agents, OpenEarth-Bench comprises 596 application cases derived from real-world research, spanning seven remote sensing application domains: urban, agriculture, vegetation, water bodies, soil, economy, and snow. Each case covers a complete EO pipeline. The statistics is illustrated in Bottom left of Fig. \ref{fig:OpenEarthbench}. In the data preparation stage, the benchmark encompasses the automatic acquisition and preprocessing of multi-source data, including RGB, multi-spectral, SAR, NTL, and remote sensing products. In the feature extraction stage, it requires agents to flexibly employ diverse methods---ranging from statistical learning and machine learning to deep learning---to accomplish the feature extraction. In the geospatial analysis stage, the benchmark involves multi-dimensional scientific computations, such as temporal trend analysis, spatial correlation analysis, and spatio-temporal coupling analysis. 

To systematically evaluate the performance of agents in open-environment EO, OpenEarth-Bench is designed with three distinctive characteristics: \textbf{1) Extensive Domain Coverage}: OpenEarth-Bench comprises 596 application cases derived from real-world research, spanning seven remote sensing domains: urban, agriculture, vegetation, water bodies, soil, economy, and snow. Detailed distribution statistics are illustrated in the bottom left of Fig. \ref{fig:OpenEarthbench}. \textbf{2) Comprehensive Pipeline Coverage:} Each case in OpenEarth-Bench encompasses a complete EO pipeline with full coverage across all stages. In the data preparation stage, the benchmark includes the autonomous acquisition and preprocessing of multi-source data, such as RGB, multi-spectral, SAR, NTL, and other remote sensing products. In the feature extraction stage, agents are required to flexibly employ diverse methodologies—ranging from statistical and machine learning to deep learning. Finally, the geospatial analysis stage involves multi-dimensional scientific computations, including temporal trend analysis, spatial correlation, and spatio-temporal coupling analysis. \textbf{3) Minimalist Tool Provision:} OpenEarth-Bench provides only large-scale remote sensing models as essential tools for specific functionalities. No auxiliary tools are provided, requiring agents to autonomously synthesize necessary tools based on the data and task context in open environments.


\begin{table}[htbp]
  \centering
  \renewcommand{\arraystretch}{1.15} 
  \caption{Comparison between OpenEarth-Bench and existing remote sensing agent benchmarks.}
  \label{tab:comparison}
  \resizebox{\textwidth}{!}{
  \begin{tabular}{lccccc}
    \toprule
    \textbf{Benchmark} & \textbf{\makecell{Data \\ Preparation}} & \textbf{\makecell{Feature \\ Extraction}} & \textbf{\makecell{Geospatial \\ Analysis}} & \textbf{\makecell{Open \\ Environment}} & \textbf{\makecell{Data \\ Modalities}} \\
    \midrule
    ThinkGeo \cite{thinkgeo} & \xmark & \cmark & \xmark & \xmark & RGB \\
    Earth-Bench \cite{earth_agent} & \xmark & \cmark & \cmark & \xmark & RGB, Multi-Spectral, Product \\
    GeoPlan-Bench \cite{earthagent} & \cmark & \cmark & \cmark & \xmark & - \\
    Cangling-Know. \cite{cangling} & \xmark & \cmark & \cmark & \xmark & RGB, Multi-Spectral, SAR \\
    \rowcolor{gray!12} 
    \textbf{OpenEarth-Bench (Ours)} & \cmark & \cmark & \cmark & \cmark & \makecell[c]{RGB, Multi-Spectral, SAR, \\ NTL, Product} \\
    \bottomrule
  \end{tabular}
  }
\end{table}

Table \ref{tab:comparison} details the comparison between OpenEarth-Bench and existing remote sensing agent benchmarks. As shown in the table, existing works are restricted by closed tool calling environments and are limited to a narrow scope of EO stages and data modalities. In contrast, OpenEarth-Bench achieves multi-source data support, full-pipeline coverage, and real-case evaluation in open environments for the first time. This provides a comprehensive and highly challenging testing platform to facilitate the development of remote sensing agents.

\subsection{Benchmark Construction and Evaluation}
To ensure the authenticity and high quality of OpenEarth-Bench in open environments, we adopt a collaborative construction paradigm consisting of ``expert guidance, code generation, and human verification''. First, senior domain experts led the definition and workflow of representative EO task cases from practical research across seven remote sensing application domains. Subsequently, leveraging advanced LLMs as an auxiliary engine, we preliminarily generated execution codes and workflow logic covering the entire ``data preparation - feature extraction - geospatial analysis'' pipeline based on the expert-defined task contexts. Finally, domain experts conducted thorough manual reviews and corrections of the model-generated codes, task performing logic, and final execution results. This process significantly improves the efficiency of benchmark construction while ensuring the scientific rigor and executability of all cases in open environments.

To facilitate evaluation in realistic open environments, \textit{OpenEarth-Bench} exclusively provides pre-trained models tailored to specific cases, omitting other predefined callable tools. This design compels the agent to adaptively plan workflows and generate tools, conditioned on the data and task context. Given the inherently long-horizon nature of EO workflows, we design our evaluation protocol to assess the agent both at individual stages and across the entire pipeline: \textbf{1) Stage-wise Evaluation:} We decouple the comprehensive EO task into three independent stages: data preparation, feature extraction, and geospatial analysis, assessing the agent's performance independently within each phase. \textbf{2) End-to-end Evaluation:} We evaluate the agent's performance across all phases as it autonomously executes the complete EO pipeline. In this setting, errors in preceding stages naturally cascade and impact subsequent operations.

To multidimensionally quantify the comprehensive efficacy of \textit{OpenEarth-Agent} in open environments, we construct an evaluation metric system encompassing effectiveness, reflective capability, and operational cost: \textbf{1) Accuracy:} Serves as the primary metric for task completion, evaluating whether the agent's output is successfully generated, correctly localized in designated storage, and strictly aligned with the ground truth. In the data preparation, we verify the consistency of metadata (e.g., data source, spatio-temporal settings) and ensure numerical discrepancies remain within a predefined error margin. In the feature extraction phase, we validate metadata alignment (e.g., value ranges, channel configurations) alongside numerical fidelity. For geospatial analysis, we confirm that the extracted insights are correctly structured as key-value pairs, with numerical deviations constrained within the acceptable tolerance. \textbf{2) Debug Rounds:} Records the number of iterations required for the agent to self-adjust and correct code based on environmental feedback when execution errors occur. This metric directly reflects the agent's capacity for self-reflection and error recovery. \textbf{3) Running Time:} Measures the total duration of the agent's planning-execution-feedback loop, evaluating the agent's operational efficiency in real-world scenarios. As cascading errors in the end-to-end evaluation can lead to disproportionate increases in debug rounds and running time for subsequent stages, we exclusively report these two metrics in the stage-wise evaluation. Accuracy is applied to both the stage-wise and end-to-end evaluations.

\section{Experiment}
\subsection{Evaluations with different LLMs on OpenEarth-Bench}

\begin{figure}[htbp]
  \centering
  \includegraphics[width=\textwidth]{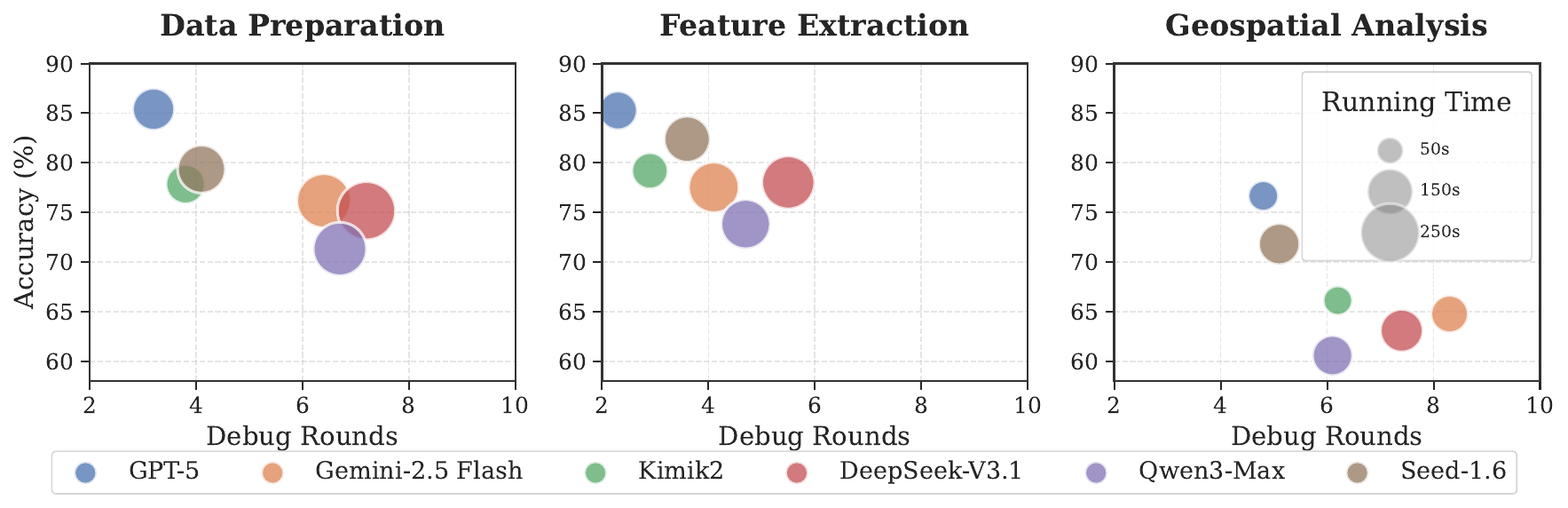}
  \caption{Results of OpenEarth-Agent based on different LLMs in Stage-Wise evaluation. The closer to the upper-left corner, the better.
  }
  \label{fig:openearth_bench_result}
\end{figure}

To evaluate the agent's performance in open environments, we benchmark OpenEarth-Agent on OpenEarth-Bench, supported by six advanced LLMs. These include two closed-source LLMs (GPT-5, Gemini-2.5-Flash \cite{gemini25}) and four open-source LLMs (Kimik2 \cite{kimi}, DeepSeek-V3.1 \cite{deepseek}, Qwen3-Max \cite{qwen}, Seed-1.6). Through an in-depth comparison of stage-wise and end-to-end evaluation results, as presented in Table~\ref{tab:openearth_bench} and Fig. \ref{fig:openearth_bench_result}, we draw the following observations:

\textit{1) Effectiveness in open environments:} Despite the diverse data and tasks in open environments, and without relying on predefined tool APIs, Table~\ref{tab:openearth_bench} demonstrates that OpenEarth-Agent achieves high accuracy across all stages in the stage-wise evaluation, with GPT-5 yielding the best performance. Furthermore, the agent leverages result-checking feedback to steadily enhance its capability. Fig. \ref{fig:openearth_bench_result} illustrates the self-correction proficiency of the evaluated LLMs, highlighting GPT-5's superior reflection and debugging abilities, which require fewer than five debug rounds across all stages.

\begin{table}[htbp]
  \centering
  \renewcommand{\arraystretch}{1.15}
  \caption{Accuracy comparison of OpenEarth-Agent powered by different LLMs on OpenEarth-Bench.}
  \label{tab:openearth_bench}
  \resizebox{\textwidth}{!}{
  \begin{tabular}{lcccccc}
    \toprule
    \multirow{2}{*}{\textbf{LLM}} & \multicolumn{3}{c}{\textbf{Stage-Wise}} & \multicolumn{3}{c}{\textbf{End-to-End}} \\
    \cmidrule(lr){2-4} \cmidrule(lr){5-7}
    & \makecell{Data \\ Preparation} & \makecell{Feature \\ Extraction} & \makecell{Geospatial \\ Analysis} & \makecell{Data \\ Preparation} & \makecell{Feature \\ Extraction} & \makecell{Geospatial \\ Analysis} \\
    \midrule
    GPT-5 & \textbf{85.40} & \textbf{85.27} & \textbf{76.66} & \textbf{82.38} & \textbf{74.83} & \textbf{58.72} \\
    Gemini-2.5-Flash & 76.17 & 77.52 & 64.76 & 74.16 & 61.58 & 45.47 \\
    Kimik2 & 77.85 & 79.19 & 66.11 & 75.84 & 64.09 & 47.81 \\
    DeepSeek-V3.1 & 75.17 & 78.02 & 63.08 & 72.82 & 60.74 & 43.12 \\
    Qwen3-Max & 71.31 & 73.82 & 60.57 & 69.79 & 56.38 & 39.26 \\
    Seed-1.6 & \uline{79.36} & \uline{82.38} & \uline{71.81} & \uline{77.18} & \uline{68.12} & \uline{51.84} \\
    \bottomrule
  \end{tabular}
  }
\end{table}

\textit{2) Complexity of cross-domain scientific analysis:} In the stage-wise evaluation, compared to the high accuracies in Data Preparation and Feature Extraction, all models exhibit a noticeable performance degradation in the Geospatial Analysis stage. The best-performing model, GPT-5, achieves an accuracy of 76.66\%, while the open-source LLMs generally hover around 60\%--70\%. This disparity reflects the fundamental difference in requirements between deep geospatial analysis and shallow data processing. While data preparation and feature extraction predominantly rely on general programming logic and mature algorithm invocations, geospatial analysis necessitates cross-domain expertise and profound analytical comprehension.

\textit{3) Impact of cascading errors in the full pipeline:} In the end-to-end evaluation, all LLMs demonstrate a baseline capability for full-pipeline execution, validating OpenEarth-Agent's potential for comprehensive cross-stage EO. However, this evaluation also reveals the issue of cascading errors inherent in long-horizon tasks. Since errors in preceding stages are progressively amplified downstream, the accuracy in the feature extraction and geospatial analysis stages drops significantly compared to the stage-wise evaluation. For instance, GPT-5's geospatial analysis accuracy sharply decreases from 76.66\% to 58.72\%, underscoring the substantial challenges agents face in end-to-end EO.

\textit{4) Execution time in real-world environments:} The runtime statistics presented in Fig. \ref{fig:openearth_bench_result} realistically reflect the agent's efficiency in open environments. All LLMs consume considerable time during the data preparation and feature extraction stages. This is primarily because data preparation involves the online retrieval and downloading of multi-source remote sensing imagery, whereas feature extraction entails online unsupervised learning and the inference of pre-trained models. Conversely, the time required for the geospatial analysis stage is significantly reduced, as it predominantly involves the analysis of structured feature data.

\subsection{Cross-benchmark evaluations on Earth-Bench}

To further validate the capabilities of OpenEarth-Agent in open environments, we conduct experiments on the cross-domain benchmark Earth-Bench \cite{earth_agent} and compare it with the tool-calling agent, Earth-Agent \cite{earth_agent}, across various LLMs. Earth-Bench comprises 104 professional tools, including pre-trained models and programmatic function tools. To ensure a fair comparison, we disabled all external knowledge and tool integration within OpenEarth-Agent. We evaluate OpenEarth-Agent under two distinct settings: 1) integrating only six essential pre-trained models from Earth-Bench to evaluate its capacity to adaptively create functionally equivalent tools, plan, and execute tasks in open environments devoid of predefined tools; and 2) integrating all available tools in Earth-Bench to assess its tool planning and invocation capabilities in a resource-rich environment. Since OpenEarth-Agent does not involve tool calls, we only use the Accuracy of this benchmark as a validation metric.

\begin{wraptable}{r}{0.6\textwidth} 
  \centering
  \renewcommand{\arraystretch}{1.15} 
  \caption{Accuracy comparison on Earth-Bench across different LLM backbones.}
  \label{tab:earth_bench}
  \begin{tabular}{lccc}
    \toprule
    \multirow{2}{*}{\textbf{LLM}} & \multirow{2}{*}{\textbf{Earth-Agent}} & \multicolumn{2}{c}{\textbf{OpenEarth-Agent}} \\
    \cmidrule(lr){3-4}
    & & \textbf{6 Tools} & \textbf{Full Tools} \\
    \midrule
    GPT-5         & 63.16 & 59.92 & \textbf{67.61} \\
    Gemini-2.5-Flash    & 55.06 & 53.44 & \textbf{57.89} \\
    Kimik2        & 56.68 & 52.63 & \textbf{59.11} \\
    DeepSeek-V3.1 & 52.23 & 51.02 & \textbf{56.68} \\
    Qwen3-Max     & 47.37 & 46.97 & \textbf{52.63} \\
    Seed-1.6      & 59.51 & 52.63 & \textbf{61.94} \\
    \bottomrule
  \end{tabular}
\end{wraptable}

The quantitative results, presented in Table~\ref{tab:earth_bench}, demonstrate that even when equipped with a minimal set of pre-trained models, OpenEarth-Agent achieves performance comparable to Earth-Agent across all evaluated LLMs. This indicates that in open environments lacking accessible tools, OpenEarth-Agent can effectively process diverse data and tasks by adaptively generating functionally equivalent tools, matching the efficacy of agents reliant on predefined tools. Furthermore, when equipped with the complete toolset, OpenEarth-Agent significantly outperforms Earth-Agent across all LLMs, highlighting its superiority over existing tool-calling agents in tool-abundant environments.


Beyond functional consistency, the tools autonomously created by OpenEarth-Agent exhibit significantly superior data adaptability compared to predefined tools. Specifically, the predefined tools in Earth-Bench are strictly tailored to the benchmark's specific data distributions. When applied to diverse data in open environments under identical task settings, they frequently encounter three critical limitations: \textbf{1) Hard-coded sensor parameters:} Numerous tools rigidify parameters for specific sensors, rendering them incompatible with alternative data sources; for instance, \texttt{compute\_tvdi} and \texttt{calculate\_ndsi} hard-code MODIS-specific scaling factors, while \texttt{lst\_single\_channel} and \texttt{temperature\_emissivity\_separation} fix specific band wavelengths. \textbf{2) Omission of invalid value masking:} Tools such as \texttt{calculate\_fvc} and \texttt{stl\_decompose} fail to process anomalous regions common in remote sensing imagery (e.g. Nodata, infinite values, cloud cover), which severely compromises downstream analytical accuracy. \textbf{3) Inappropriate numerical processing:} Several tools apply arbitrary numerical operations that corrupt the intrinsic physical significance of the data. For example, \texttt{calculate\_tif\_average} and \texttt{calc\_batch\_image\_mean\_threshold} apply naive min-max linear stretching to physical metrics like NDVI and LST, destroying their absolute physical meanings. In contrast, OpenEarth-Agent overcomes these bottlenecks by leveraging active data perception to dynamically acquire real-time metadata alongside spatio-temporal-modal characteristics. This holistic data comprehension enables the agent to adaptively synthesize highly versatile, robust tools tailored to the specific data context, ensuring robust generalization unconstrained by the limitations of restricted data scenarios.



\subsection{Ablation Study}

To validate the effectiveness of the multi-agent collaborative architecture and the open knowledge and tool integration modules, we conduct a comprehensive ablation study on the GPT-5-powered OpenEarth-Agent using the OpenEarth-Bench.


\begin{table}[htbp]
  \centering
  \renewcommand{\arraystretch}{1.15}
  \caption{Ablation study on the multi-agent architecture.}
  \label{tab:multi_agent_ablation}
  \resizebox{\textwidth}{!}{
  \begin{tabular}{ccc ccc ccc}
    \toprule
    \multirow{2}{*}{\textbf{\makecell{Data \\ Summary}}} & \multirow{2}{*}{\textbf{\makecell{Plan \& \\ Workflow}}} & \multirow{2}{*}{\textbf{\makecell{Result \\ Check}}} & \multicolumn{3}{c}{\textbf{Stage-Wise}} & \multicolumn{3}{c}{\textbf{End-to-End}} \\
    \cmidrule(lr){4-6} \cmidrule(lr){7-9}
    & & & \makecell{Data \\ Preparation} & \makecell{Feature \\ Extraction} & \makecell{Geospatial \\ Analysis} & \makecell{Data \\ Preparation} & \makecell{Feature \\ Extraction} & \makecell{Geospatial \\ Analysis} \\
    \midrule
    \xmark & \cmark & \cmark & 84.56 & 77.85 & 67.62 & 81.71 & 74.16 & 57.88 \\
    \cmark & \xmark & \cmark & 83.72 & 82.21 & 71.81 & 81.54 & 69.13 & 50.67 \\
    \cmark & \cmark & \xmark & 78.02 & 79.86 & 65.77 & 75.84 & 59.73 & 34.73 \\
    \rowcolor{gray!12} 
    \cmark & \cmark & \cmark & \textbf{85.40} & \textbf{85.27} & \textbf{76.66} & \textbf{82.38} & \textbf{74.83} & \textbf{58.72} \\
    \bottomrule
  \end{tabular}
  }
\end{table}

\textbf{Effectiveness of the Multi-Agent Architecture.} While a single code-based agent is capable of creating tools and executing tasks through code generation, we conducted ablation studies on other specialized agents to investigate the advantages of multi-agent architecture over a single coding agent. We ablate the Data Summary, Plan \& Workflow, and Result Check agents, respectively. The results, reported in Table \ref{tab:multi_agent_ablation}, reveal the indispensable roles of each agent in handling long-horizon tasks within open environments:

\textit{1) Data Summary Agent:} In the Stage-Wise mode, removing this module leads to a substantial accuracy drop in the feature extraction and geospatial analysis stages. This decline occurs because the agent encounters unknown input data in these mid-to-late stages; lacking a summary module deprives the system of its capability to perceive data characteristics. Notably, its absence has a relatively minor impact during End-to-End evaluations. This demonstrates that in full-pipeline tasks, metadata acquired during the data preparation phase can flow efficiently within the agent system.

\textit{2) Plan \& Workflow Agent:} The absence of this module triggers severe performance degradation, particularly in the End-to-End long-horizon evaluation, where the accuracy of the terminal geospatial analysis stage plummets to 50.67\% (an approximate 8\% decrease). This strongly indicates that in open environments, agents lacking adaptive planning are highly susceptible to severe ``cascading errors'' when confronting highly complex EO tasks.

\textit{3) Result Check Agent:} Removing this module forces the OpenEarth-Agent into a one-shot execution paradigm. Experiments show that without environmental feedback and trial-and-error opportunities, the system experiences a drastic accuracy drop across all stages. This underscores that real-time self-correction mechanisms are essential prerequisites for ensuring high task success rates in open remote sensing environments.


\begin{table}[htbp]
  \centering
  \renewcommand{\arraystretch}{1.15}
  \caption{Ablation study on open knowledge and tool integration.}
  \label{tab:knowledge_tool_ablation}
  \resizebox{\textwidth}{!}{
  \begin{tabular}{cc ccc ccc}
    \toprule
    \multirow{2}{*}{\textbf{\makecell{Knowledge \\ Integration}}} & \multirow{2}{*}{\textbf{\makecell{Tool \\ Integration}}} & \multicolumn{3}{c}{\textbf{Stage-Wise}} & \multicolumn{3}{c}{\textbf{End-to-End}} \\
    \cmidrule(lr){3-5} \cmidrule(lr){6-8}
    & & \makecell{Data \\ Preparation} & \makecell{Feature \\ Extraction} & \makecell{Geospatial \\ Analysis} & \makecell{Data \\ Preparation} & \makecell{Feature \\ Extraction} & \makecell{Geospatial \\ Analysis} \\
    \midrule
    \xmark & \xmark & 79.69 & 81.21 & 72.48 & 78.02 & 66.78 & 47.65 \\
    \xmark & \cmark & 80.87 & 84.40 & 73.82 & 78.86 & 69.46 & 51.34 \\
    \cmark & \xmark & 84.56 & 82.04 & 74.16 & 81.20 & 70.80 & 52.68 \\
    \rowcolor{gray!12} 
    \cmark & \cmark & \textbf{85.40} & \textbf{85.27} & \textbf{76.66} & \textbf{82.38} & \textbf{74.83} & \textbf{58.72} \\
    \bottomrule
  \end{tabular}
  }
\end{table}

\textbf{Effectiveness of Knowledge and Tool Integration.} To isolate and quantify the performance gains yielded by the external knowledge and tool bases, we perform corresponding ablation experiments. The results in Table \ref{tab:knowledge_tool_ablation} demonstrate their crucial and complementary roles:

\textit{1) Knowledge Integration:} This integration yields the most significant improvements in the data preparation and geospatial analysis stages. During the data acquisition phase, cross-domain knowledge guides the agent to accurately match the required multi-source heterogeneous data. During the analysis phase, domain knowledge provides the agent with indispensable geoscience priors, ensuring the scientific validity of the analytical logic.

\textit{2) Multi-Stage Tool Integration:} Tool integration predominantly enhances the feature extraction and geospatial analysis stages. When facing feature extraction and complex analytical demands in open environments, the extensive tool repository offers abundant algorithmic references, safeguarding the correctness and robustness of the tools generated by OpenEarth-Agent.

\textit{3) Synergistic Effect:} When both modules are enabled, the model achieves optimal performance across all metrics, indicating that the integration of knowledge and tools exerts a complementary effect, jointly elevating the agent's overall capabilities.

\section{Discussion}

\subsection{Limitations and Expectations}

While OpenEarth-Agent demonstrates strong generalization capabilities in open environments, its dynamic tool creation paradigm fundamentally depends on the underlying LLM's coding and reasoning proficiency. Although code-probing and iterative feedback mechanisms mitigate many errors, the agent may occasionally synthesize tools that are syntactically correct but functionally suboptimal for highly complex geospatial modeling. Furthermore, the continuous cycle of real-time data perception, dynamic DAG planning, and iterative tool refinement requires multiple LLM inference calls. This introduces higher computational overhead and processing latency compared to traditional, static predefined tool-calling pipelines, which may currently limit its deployment in highly time-critical emergency response scenarios.

Moving forward, our future work will focus on two primary expectations. First, we aim to optimize inference efficiency by implementing a robust tool-caching mechanism, allowing successfully synthesized and verified tools to be archived and reused as templates to reduce redundant LLM API calls. Second, we plan to directly integrate physical laws and geospatial topological constraints into the tool synthesis prompts, which will strictly bound the agent's action space and further reduce logical hallucinations in complex Earth Observation tasks.

\subsection{Societal Implications and Environmental Considerations}

The deployment and operation of OpenEarth-Agent, particularly its reliance on continuous LLM inference for dynamic tool creation and iterative feedback, necessitate substantial computational resources. This intensive computational pipeline inherently leads to significant energy consumption, constituting a critical environmental concern that contributes materially to the carbon footprint and associated ecological impacts. While the framework democratizes complex Earth Observation analysis and aids in monitoring environmental changes, we must critically recognize these externalities.

Adopting sustainable energy solutions can demonstrably reduce the ecological ramifications of autonomous geospatial agents, aligning technological advancement with environmental stewardship. It is incumbent upon the research community to proactively evaluate and address the environmental costs inherent in deploying computationally demanding agentic methodologies. Key mitigation approaches must encompass not only sustainable energy sourcing but also the continuous pursuit of algorithmic optimization—such as implementing efficient tool-caching to minimize redundant generation—and the utilization of energy-efficient, distilled models to reduce overall computational overhead. 

\section{Conclusion}
\label{sec:conclusion}
In this paper, we introduce OpenEarth-Agent, the first tool-creation agent framework tailored for open-environment EO. By shifting the paradigm toward adaptive workflow planning and tool creation conditioned on data and task contexts, and coupling this with open-ended multi-stage tool and cross-domain knowledge integration, OpenEarth-Agent successfully executes end-to-end EO pipelines across multiple domains. Furthermore, to provide a rigorous evaluation platform, we construct OpenEarth-Bench, the first open-environment benchmark for remote sensing agents, comprising $596$ full-pipeline application cases across 7 application domains. Extensive evaluations on OpenEarth-Bench, alongside cross-benchmark experiments on Earth-Bench, firmly validate the efficacy of our proposed framework. Notably, it achieves competitive performance using only a minimal set of essential tools compared to baselines relying on full tool calling, and outperforms existing agents when provided with the complete toolset. Ultimately, OpenEarth-Agent establishes a robust and generalizable foundation for the future of autonomous EO in open environments.

\clearpage
{\small

}
\clearpage

\end{document}